\documentclass[conference]{IEEEtran}
\IEEEoverridecommandlockouts
\usepackage{cite}
\usepackage{amsmath,amssymb,amsfonts}
\usepackage{algorithmic}
\usepackage{graphicx}
\usepackage{textcomp}
\usepackage{xcolor}
\usepackage{graphicx}
\usepackage{subfig}
\usepackage{enumitem}
\usepackage{amsmath}
\usepackage{adjustbox}
\usepackage{hyperref}
\def\BibTeX{{\rm B\kern-.05em{\sc i\kern-.025em b}\kern-.08em
    T\kern-.1667em\lower.7ex\hbox{E}\kern-.125emX}}

\begin{document}

\newcommand{\nb}[2]{
    \fbox{\bfseries\sffamily\scriptsize#1}
    {\sf\small\textcolor{red}{\textit{#2}}}
}
\newcommand\ag[1]{\nb{AG}{#1}}
\newcommand\tp[1]{\nb{TP}{#1}}

\title{Explainable Fundus Image Curation and Lesion Detection in Diabetic Retinopathy}

\author{\IEEEauthorblockN{Anca Mihai and Adrian Groza}
\IEEEauthorblockA{
\textit{Artificial Intelligence Research Institute AIRi@UTCN}\\
\textit{Department of Computer Science} \\
\textit{Technical University of Cluj-Napoca, Cluj-Napoca, Romania} \\
Mihai.Ge.Anca@student.utcluj.ro, Adrian.Groza@cs.utcluj.ro}
}

\maketitle
\begin{abstract}
Diabetic Retinopathy (DR) 
affects individuals with long-term diabetes. 
Without early diagnosis, DR can lead to vision loss. 
Fundus photography 
captures the structure of the retina 
along with abnormalities indicative of the stage of the disease. 
Artificial Intelligence (AI) can support clinicians in identifying these lesions, reducing manual workload, but models require high-quality annotated datasets. 
Due to the complexity of retinal structures, errors in image acquisition and lesion interpretation of manual annotators can occur. We proposed a quality-control framework, ensuring only high-standard data is used for evaluation and AI training. First, an explainable feature-based classifier is used to filter inadequate images. The features are extracted both using  image processing and contrastive learning. Then, the images are enhanced and put subject to annotation, using deep-learning-based assistance. Lastly, the agreement between annotators calculated using derived formulas determines the usability of the annotations.
\end{abstract}

\vspace{0.3cm}

\section{Motivation}
\begin{figure}
    \centering
    \includegraphics[width=0.48\textwidth]{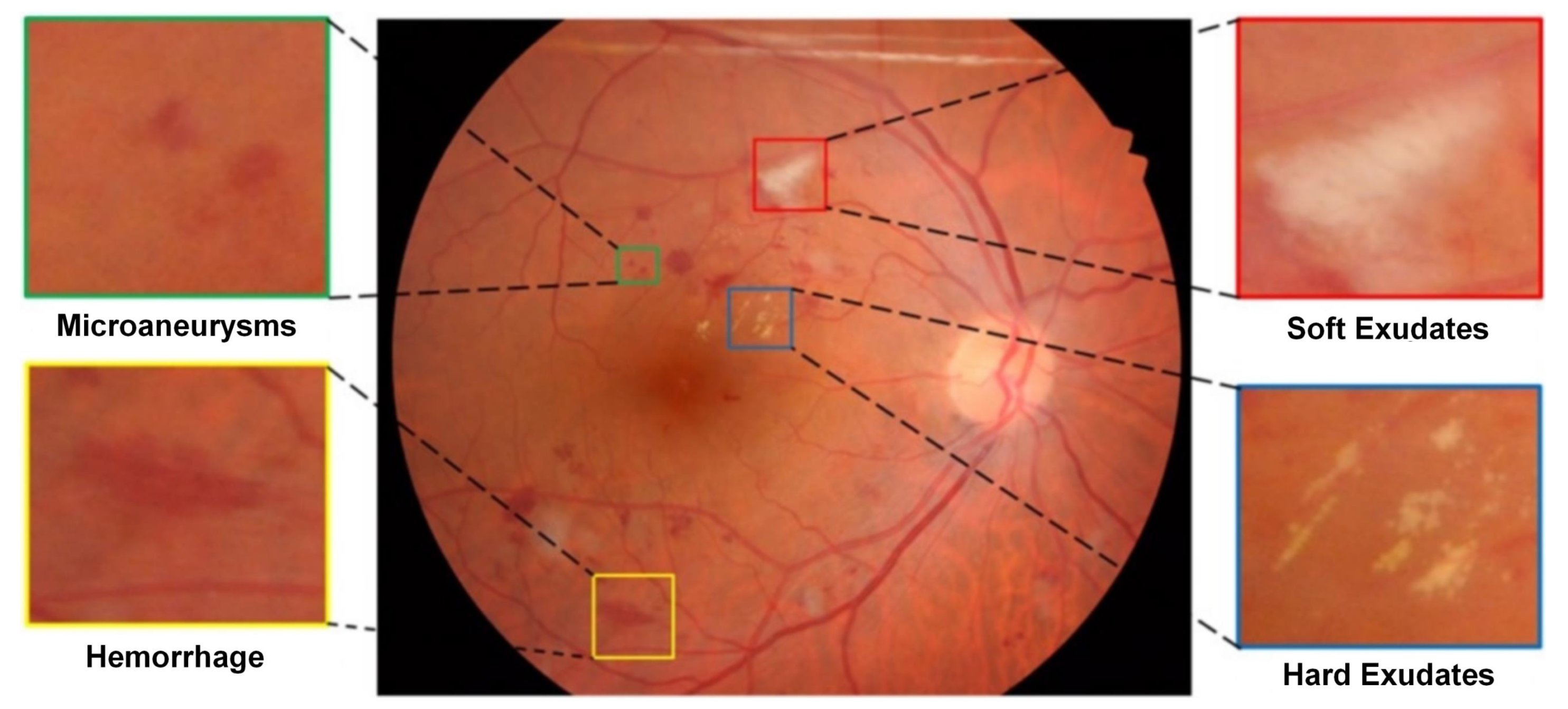}
    \caption{Different lesions in fundus images}
    \label{figlesion}
\end{figure}

In ophthalmology, fundus imaging is widely used for evaluating retinal health and for the diagnosis and monitoring of retinal diseases such as diabetic retinopathy (DR). DR manifests through characteristic retinal abnormalities (Figure~\ref{figlesion}), including red lesions-microaneurysms (MA) and intraretinal hemorrhages (HA)-as well as yellow-white lesions such as hard exudates (EX) and whitish lesions such as soft exudates/cotton-wool spots (SE).


DR can be non-proliferative diabetic retinopathy (NPDR) or proliferative diabetic retinopathy (PDR). 
The evolution NPDR can be further divided into more stages~\cite{Stages}:
(i) Mild NPDR, only MAs are present; 
(ii) Moderate NPDR,  HAs or MAs in 1-3 of 4 regions of the image and/or SEs, EXs or venous beading; 
(iii) Severe NPDR, $>$20 HAs in each one of the 4 regions, venous beading in 2 or more regions of 4 and intraretinal microvascular abnormality in at least 1 quadrant. 
Diabetic Macular Edema (DME) indicates worsening of DR, appearing from \textit{Moderate NDPR} stage onwards,  
characterised by the presence of excess fluid in the proximity of the macula of the retina.
EXs that are within 500 $\mu\text{m}$ of the center of the macula (fovea) can suggest the presence of clinically significant Macular Edema (CSME), associated with DME.

Manual interpretation of these findings in fundus images is time-consuming and analysis offered by AI can support clinicians in identifying abnormalities, reducing manual workload, and therefore enabling faster, more efficient screening. 
However, AI models require well-annotated datasets to achieve high accuracy. 
For precise segmentation, pixel-level annotations are required. 


However, both fundus imagery and its associated annotation process entail several challenges. First, the complexity of retinal anatomy can lead to acquisition artifacts and variability; consequently, degraded image quality may mask clinically relevant features and increase the likelihood of misinterpretation. 
Moreover, publicly available datasets with pixel-level annotations are highly curated and relatively limited. As noted in~\cite{ImageAcquisition}, although models trained on such datasets can achieve high accuracy, their performance and reliability on real-world clinical images remain difficult to ascertain.

Second, poor-quality images may compromise the accuracy of manual annotations themselves. In addition, the available online labelling tools are either offering only polygon-based annotations or hard-to-use pixel-level options, as lesions in fundus images are small.

Third, dependence on a single manual annotator is a potential source of bias affecting data quality, especially in the context of fundus images, where lesions are hard to detect and differentiate from each other~\cite{Distiction}.

To mitigate these limitations, we propose a labeling framework for the curation of DR fundus images and their corresponding annotations.
First, image quality is evaluated using an explainable and robust classifier that combines handcrafted image-processing descriptors with features learned via contrastive learning. Images deemed of sufficient quality then undergo lesion-visibility enhancement using Contrast Limited Adaptive Histogram Equalization (CLAHE), followed by gamma correction to regulate brightness. During manual labeling, the annotation interface provides deep-learning–based lesion suggestions to support clinicians and facilitate lesion identification. Finally, the usability and consistency of the curated images and annotations are assessed through inter-annotator agreement measures, enabling bias reduction and improved labeling reliability.



\section{Related Work}


Methods for assessing fundus image quality are generally grouped into two categories: approaches based on manually engineered features and learning-based techniques. Paulus et al.~\cite{Haralick} proposed quality criteria comprising (i) structural aspects—such as the visibility of the optic disc and the retinal vasculature - and (ii) generic aspects, including homogeneous illumination, adequate brightness, and high contrast. To operationalize these criteria, they employed three metrics: (i) K-means clustering to quantify the separability of anatomical structures, (ii) a sharpness measure, and (iii) Haralick texture descriptors. The extracted features can be aggregated into a composite quality score for overall assessment; for instance, the computed metrics may be concatenated into a feature vector and classified using a Support Vector Machine. 
Using cross-validation for evaluation, the proposed system achieved an accuracy of 0.867.
 
Zago et al.~\cite{QAInceptionv3} employed a pretrained InceptionV3 convolutional neural network and fine-tuned it for fundus image quality assessment, achieving an AUC of approximately 99\%. Khalid et al.~\cite{XAIDeepLearning} addressed the interpretability challenges of deep learning models by surveying a range of explainability techniques, including post-hoc approaches (e.g., gradient/saliency maps and occlusion-based analyses) and intrinsic methods (e.g., concept whitening). However, despite highlighting influential regions, these methods typically do not elucidate why particular areas affect the model’s decision. In particular, when a region drives a low quality score, they often cannot specify the precise degradation (e.g., blur, illumination artifacts, or occlusion) responsible for the poor assessment.
 
Yi et al.~\cite{QAClip} have relied on Contrastive Language-Image Pre-training (CLIP) model with zero-shot transfer learning. 
For assessing the quality of medical images, antonym prompts ("Good photo" versus "Bad photo", "Bright photo" versus "Dark photo") were considered. 
To tackle the challenge that CLIP works with fixed-sized inputs, positional embeddings of the Transformer model are replaced with ResNet. The method reaches 0.95 accuracy and 0.89 F1 score.


For extracting lesions, deep learning methods have been widely used as they prove accuracy and time efficiency at inference. 
For segmenting EXs, Prentasic et al.~\cite{Exudates} have used Convolutional Neural Networks (CNNs). 
A workflow of three steps: (i) augmentation by rotating and upsampling the training set, 
(ii) removal of optical disk, (iii) extraction of green channel - has lead to sensitivity of 0.77. 

For extraction of HAs and MAs,  Khojasteh et al.~\cite{HAMA} have proposed extraction of patches of images previously having their contrast enhanced. 
The obtained parts are fed into CNN and probability maps for each of the lesion types considered are used. 
The isolated pixels and spread due to convolution are automatically removed. Specificity scores of 0.84 for HAs and 0.85 for MAs are obtained.

Boulaabi et al.~\cite{DeepLabV3} have highlighted the performance of \textit{DeepLabV3+} architecture, obtaining 0.99 accuracy for all four types of lesions. 
The images are first cropped and enhanced using contrast-limited adaptive histogram equalization on L channel of LAB image space. 
In addition, due to the limited dataset, augmentation is used. 
At the same time, Orolando et al.~\cite{Hybrid} have selected possible candidates using the green channel of the image on which r-polynomial transformation, followed by Gaussian filer is applied. Features extracted \textit{both from a CNN and manually}
(which are based on the color bands obtained after illumination correction and contrast enhancement) are then put altogether. A Random Forest generates a probability map based on the feature vector, resulting in a 0.49 per lesion sensitivity.

\section{Datasets Used}

There are public datasets in which image quality has been explicitly labeled:
DRIMDB containing 125 high quality images and 69 low quality images, 
DeepDRID~\cite{DeepDRID} with 387 images or~\cite{BlurryDataset} containing 160 blurry images.
In addition, Indian Diabetic Retinopathy Image Dataset (IDRiD) is composed of 516 high-quality images, including for 87 images pixel-level annotations for Microaneurysms, Haemorrhages, Hard Exudates, Soft Exudates and Optic Disc.

To ensure generalization and robustness, DRIMDB, DeepDRID, \cite{BlurryDataset} and "Disease Grading" subset of IDRID are fused to achieve a moderately balanced distribution of both good and bad images, resulting in approximately 1300 images: 441 are graded as "bad", while 806 as "good".

Regarding lesion detection, even though \textit{IDRiD} dataset contains high quality annotations, there is a limited number of images, taken under similar lighting and acquisition conditions. Therefore, even though models trained on this dataset, including~\cite{DeepLabV3} demonstrated very good results, it does not guarantee whether they would generalize well to real-world clinical images. Similar to IDRiD, DIARETDB1 is a curated dataset composed of 89 fundus images with annotations for HAs, Red Dots (MAs) and EXs, and thus can not reflect the variability found in real-world fundus images. \textit{e-Ophta} shares an analogous problem: it consists of high quality annotations for HAs, but the dataset contains only 47 images.

To address the limitations of the datasets mentioned above and ensure the framework is robust, \textit{Retinal Lesions Dataset} \cite{RetinalLesionsDataset} is used in the training and testing of the segmentation models: it consists of 1.500 images with all four types of lesions. 
Although it is more general, the dataset presents certain issues: the annotations are enlarged and in some cases HAs and MAs are annotated within overlapping regions. Despite these challenges, it is well-suited for providing lesion recommendations, aligning with the goal of the application. Moreover, it highlights even more the importance of the framework: the necessity for inter-rater reliability assessment and curation.

\section{Fundus Image Quality Assessment}


\subsection{Feature-based classification}


For classification tasks, linear models represent the most transparent methods, followed by \textit{decision trees} and \textit{random forests}, which offer a deepsight into the decision-making process. 
Even though deep learning models have proved high accuracy in grading fundus images, they offer little insight into how decisions are made. 
Therefore, for quality assessment, six of manually engineered features are extracted: 
(i) brightness; 
(ii) top-hat transform, followed by Frangi filter;
(iii) BRISQUE;
(iv) Euclidean norm on Sobel Operators;
(v) entropy, 
(vi) Peak-to-Mean.


First, an image can be assessed whether it is underexposed or overexposed by calculating the \textit{Brightness} (i.e. average of pixel values): if it is too low it is underexposed, if too high it is overexposed; fundus images have a darker background, meaning that the threshold for underexposure is usually very small (below 30-70 on a 0-255 scale), whereas for overexposure the threshold is around 160-200. 
For determining the brightness, the images are first cropped to exclude the black margins which significantly affect the score. 

Second, for structural assessment, fundus images should capture key structures as vessels. \textit{Top-hat transform} is used to highlight the fine details on image\cite{TopHat}, 
followed by application of \textit{Frangi filter} for accentuating tubular-like artifacts, including vessels \cite{Frangi}. 

Third, some metrics are based on how "naturalness" is perceived by people~\cite{Non-Reference}. \textit{BRISQUE} (Blur/Repeat/Independent Que) divides the images, analyzes local statistical features of each piece and predicts the image quality based on the learned differences. Lower scores reflect better quality of image.

Fourth, one of the other key factors for determining the quality represents the \textit{Sharpness}, 
blurry images making it difficult to extract important regions like vessels, lesions, etc. To evaluate the blurriness of an image, \textit{gradient and variants} can be used: the derivatives with regard to the two dimensions of an image can be computed and combined. 
Paulus et al.~\cite{Haralick} have used the Euclidean norm as a sharpness metric, where higher scores indicate sharper images. This approach will also be used in this paper, being applied on \textit{Sobel operators}, which determine the difference between pixels around an edge: the higher the difference is, the sharper the edge.

Fifth, \textit{Entropy} measures the information content of the pixel intensity distribution. 
In fundus images, higher entropy reflects greater detail and variation, whereas lower entropy indicates uniformity, which is sometimes undesirable due to the loss of important structural features.

Sixth, a high \textit{Peak-to-Mean} indicates the presence of strong peaks or outliers relative to the average level, which sometimes determines the \textit{sharpness} or \textit{level of contrast}.


The extracted features serve as input to a binary classification model.

\subsection{Usage of Visual Language Models}


Determining Sharpness through Sobel operators, along with computationally intensive Frangi filter and BRISQUE, are very time-consuming, making them less ideal for clinical settings where real-time processing is crucial.


Concurrently, Vision–Language Models (VLMs) integrate visual encoders with large language models to learn a shared multimodal representation. In typical architectures, an image is mapped to a sequence of embeddings via a vision backbone (e.g., a ViT or CNN), which are then aligned with text embeddings through contrastive pretraining (e.g., CLIP-style objectives) or fused via cross-attention in a multimodal transformer. This alignment enables the model to associate visual patterns with natural-language concepts and to produce explanations or descriptions that connect pixel-level evidence to semantic interpretations. A key capability enabled by such pretraining is zero-shot learning: given a set of candidate class descriptions (prompts), the model can infer labels for previously unseen categories by comparing image and text representations (or by using the language model to reason over the visual tokens) without task-specific retraining. This property makes VLMs flexible and potentially more generalizable than closed-set classifiers, suggesting their suitability for fundus image interpretation, where variations in acquisition conditions and lesion appearance are common.

CLIP is leveraged by using contrastive text inputs. It computes embeddings for the image and available text labels. Then, by comparing these embeddings, through cosine similarity, it measures how well the image matches each of the text description options.

Sharpness is assessed by using the following contrasting text prompts: "This photo is blurry and lacks focus." versus "This photo is sharp and highly detailed.". The probability of the image matching with the first description is the feature, emphasizing the  contrast feature, "Blurriness" of the image.

For assessing the structural quality score, the contrastive prompts are: "A blurry fundus photograph with unclear anatomical landmarks." in opposition with "A sharp fundus image with clearly visible optic disc and blood vessels". 
Although sharpness is explicitly considered in the previous feature, similar terminology ("blurry" vs. "sharp") is employed here to emphasize structural degradations, such as loss of anatomical detail. This design helps the contrastive model capture structural artifacts more effectively. The probability of the image matching with the first input, wrapped in a feature called "Artifacts" is used in the binary classification.

\section{Fundus Image Lesion Dectection and Enhancement}
Manual annotation of large, spatially fragmented lesions - such as hard exudates (EX) - is often time-consuming for ophthalmology experts and can be impractical when high boundary precision is required. Clinicians indicated that presenting automatically generated lesion regions, followed by manual verification and refinement, would be advantageous~\cite{bilc2021interleaving}. 
Accordingly, we incorporate a lesion prediction module into the proposed framework to support annotation and reduce the workload associated with fully manual labeling.


\begin{figure}
  \centering
  \subfloat[Initial image]{%
    \includegraphics[width=0.22\textwidth, height=5cm]{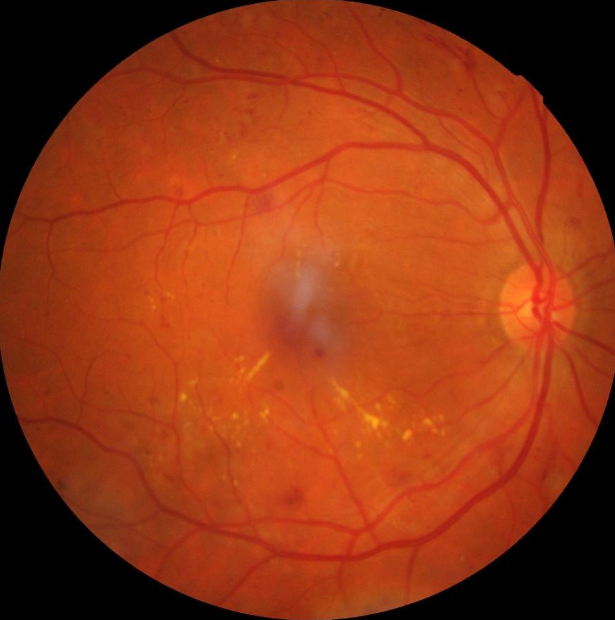}
    \label{fig:image1}
  }\hfill
  \subfloat[Enhanced image]{%
    \includegraphics[width=0.22\textwidth, height=5cm]{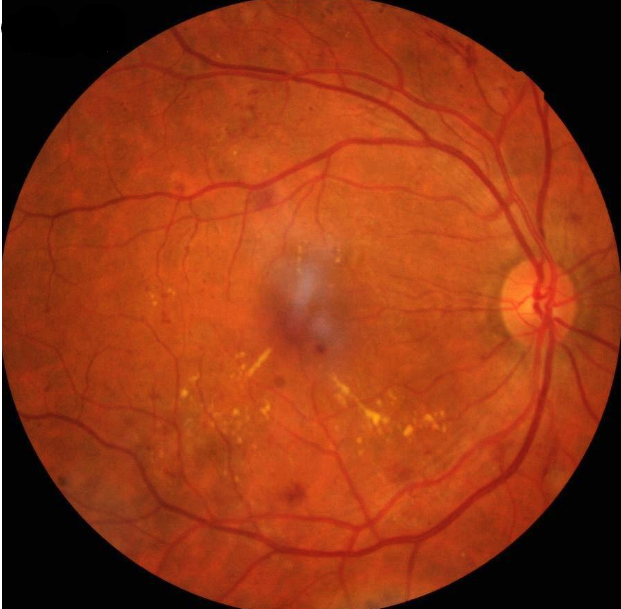}
    \label{fig:image2}
  }
  \caption{Effects of enhancement of images}
  \label{fig:sidebyside}
\end{figure}

\subsection{Quality Enhancement of Fundus Images}

To facilitate the visualization of small lesions, enhancement methods are applied to improve the contrast and highlight the abnormalities both in bright and dark regions.
Two pixel-based techniques are used~\cite{Non-Reference}:
First, \textit{histogram equalization} helps in the case of underexposed or overexposed images; nevertheless, it may eliminate details and that is why improvements have been proposed: \textit{Contrast-Limited Adaptive Histogram Equalization} aims to achieve local contrast enhancement.
Second, \textit{gamma correction} aims to compensate for the potential loss in brightness determined by the application of CLAHE. 
Figure~\ref{fig:sidebyside} shows the effects of the enhancement, making subtle lesions more prominent and defined. 

\subsection{Training lession-specific models}

To reduce the likelihood of missing small lesions, particularly microaneurysms (MAs), we train individual lesion-specific models. 
While Wei et al.~\cite{RetinalLesionsDataset} adopted a patch-based strategy for detecting multiple lesion types (HEs, HAs, and EXs), our approach applies whole-image segmentation for hemorrhages and exudates to preserve global anatomical context and improve computational efficiency, since exhaustive sliding-window inference can be costly. In contrast, MAs are modeled using patch-based training, as their minute spatial extent is often insufficiently represented at full-image resolution. 

The dataset is split into training, validation and test parts using a 70-15-15 ratio. 
To improve model generalization and simulate diverse imagining conditions, data augmentation is applied on the training set: random scaling, slight rotation, brightness/contrast adjustment, blurring and gaussian noise inclusion. Normalization and resizing is applied to all parts.

Multiple architectures are explored, including \textit{DeepLabV3+}, \textit{DeepLabV3} and \textit{Unet++}. ResNet-101 is used as the backbone network, providing more depth than ResNet-50, allowing the model to capture more complex feature and contextual information.

We investigate two loss functions. The first is a composite objective defined as the mean of Dice loss and binary cross-entropy (BCE) loss. Dice loss is well suited to segmentation under severe class imbalance, as is typical in fundus images where lesions such as microaneurysms occupy only a small fraction of the field of view. In contrast, BCE penalizes errors at the pixel level, promoting accurate local classification. Their combination therefore balances region-level overlap with pixel-wise fidelity, which is particularly advantageous when optimizing against detailed (and potentially enlarged) lesion annotations.
The second objective is the focal Tversky loss, which further mitigates class imbalance by weighting false positives and false negatives asymmetrically and focusing learning on harder examples. To reduce overfitting, early stopping is employed during training.


\subsection{Post-processing of predicted annotations}
With the exception of SEs, most lesion types are small and sparsely distributed. Consequently, blob-like automated segmentations often include portions of surrounding background. To streamline subsequent manual review and correction, post-processing can be applied to suppress spurious regions and retain only the most plausible lesion candidates.


For hard exudates (HEs), candidate pixels are first identified based on intensity using the V channel in the HSV color space. Their characteristic coloration - ranging from whitish to yellowish - is then assessed via complementary constraints on the S and H channels. To improve discrimination from surrounding retinal tissue, these features are evaluated both in absolute terms and relative to the local neighborhood (e.g., via local contrast or deviations from nearby statistics).


For hemorrhages (HAs) and microaneurysms (MAs), candidate regions are detected as areas with reduced intensity relative to their local surroundings, primarily using the V channel in HSV space. Spurious detections and small artifacts are subsequently suppressed via morphological opening to remove insignificant regions and retain coherent lesion candidates.


\section{Inter-annotator agreement}

Inter-expert variability - driven by differences in experience, fatigue, and domain expertise~\cite{Expertize} - can introduce annotation noise that may substantially affect the performance and generalization of AI models~\cite{Annotator}. Consequently, inter-annotator agreement metrics are commonly employed to quantify the level of consistency among evaluators and to assess the reliability of the resulting labels.

The protocol has three steps:
First, for each image, pair-wise agreements between the existing annotations are calculated: the average of scores per type of lesion is calculated.
Second, the "outlier" annotations can affect the score and are detected by determining the number of other annotations with which the agreement is low: if it surpasses a certain threshold, the annotations of the respective clinician are discarded. 
Third, an overall agreement of remaining annotations for a certain image is determined: if it is too low, the image together with the annotations are discarded.


For Pair-Wise Agreements, the Cohen's Kappa Coefficient ranges from -1 to 1, where 0.41– 0.60 signifies moderate agreement, 0.61–0.80  substantial agreement, 0.81–1.00 as almost perfect agreement, and -1 indicates complete disagreement.
Usually it is used for categorical annotations, but according to Yang et al.\cite{IAA}, it can be adapted to images, considering each pixel as a subject for agreement:
\begin{equation}
  K= \frac{\bar{P} - \bar{P}_e}{1 - \bar{P}_e},
\end{equation}
where:
\begin{equation}
 \bar{P} = \frac{(A + D)}{A + B + C + D},
\end{equation}
\begin{equation}
    \bar{P}_e = \frac{(A + B)(A + C)}{(A + B + C + D)^2} + \frac{(C + D)(B + D)}{(A + B + C + D)^2}.
\end{equation}   

Here, 
A denotes pixels for which both annotators assign the foreground class; 
B and 
C denote pixels labeled as foreground by only one of the annotators; and 
D denotes pixels for which both annotators assign the background class.


The formula has been adapted to consider expertize and confidence, both with levels in the range 0-1:
  \begin{eqnarray}
  A = \sum_{i,j,p} ((p_i > 0) \land  (p_j > 0)) \cdot p_i \cdot p_j
  \label{eq:sumA}\\
  B = \sum_{i,j,p} ((p_i > 0) \land  (p_j = 0)) \cdot p_i
  \label{eq:sumB}\\
  C = \sum_{i,j,p} ((p_i = 0) \land  (p_j > 0)) \cdot p_j
  \label{eq:sumC}\\
D = \sum_{i,j,p} (p_i = 0) \land  (p_j = 0)
  \label{eq:sumD}
\end{eqnarray}
where $i$ and $j$ are annotations and $p_i = confidence_i \times expertise_i$ represents the value of a pixel given the confidence with which the lesion is marked and expertise of the annotator.

The values for confidence and expertise level are:
\begin{equation}
\text{confidence} \in
\begin{cases}
[.0, .2) & \text{very low} \\
[.2, .4) & \text{low} \\
[.4, .6) & \text{medium} \\
[.6, .8) & \text{high} \\
[.8, 1] & \text{very high}
\end{cases}
\label{eq:confidence_range1}
\end{equation}

\begin{equation}
\text{expertise} \in
\begin{cases}
[.0, .1) & \text{No medical background} \\
[.1, .3) & \text{Medical student} \\
[.3, .5) & \text{Doctor in another specialty} \\
[.5, .9) & \text{Resident/junior ophthalmologist} \\
[.9, 1] & \text{Expert ophthalmologist}
\end{cases}
\label{eq:confidence_range}
\end{equation}

   
The Dice Similarity Coefficient with values in 0-1 can be weighted with confidence and expertise of the annotator.




  

  
  



\section{Results}

\subsection{Quality Assessment}

Multiple classifiers have been tested on the combined dataset, classifying images as either "good" or "bad".
Data is split into 70-30\% for training/test parts to prevent overfitting. 
The results provided in Table~\ref{tab:classifier} are on the test set, using the classic image processing features. Here, one can see that Random Forest classifier achieves the best performance considering both F2 Score and accuracy, followed by the XGBoost classifier and Logistic Regression.

Additionally, a grid search was conducted to optimize the hyperparameters of each classifier and to handle potential imbalance-related issues that might arise after splitting the data. 
The F2 score was chosen to prioritize minimizing false negatives, thereby ensuring that high-quality images are not discarded.
Using contrastive learning features and excluding BRISQUE further improves the overall results (see Table~\ref{tab:classifier}), where the most accurate two classifiers from previous table are being used. 
\begin{table}
\centering
\caption{Features-based vs. VLM binary classification}
\begin{tabular}{llcc}
& Classifier & F2 Score & Accuracy \\
\hline
Featured based \\ \hline
\textit{Brightness,}  & Random Forest & .90 & .82  \\
\textit{Top-hat+Frangi,} & Logistic Regression & .85 &  .76 \\
\textit{Sobel+Euclidean Norm,}  &  XGBClassfier &.92 & .74 \\
\textit{BRISQUE,Entropy,} & & & \\
\textit{Peak to Mean} & & & \\
\hline
Visual Language Model \\ \hline
\textit{Brightness, Blurriness,} & Random Forest & .93 & .86  \\
\textit{Absence of Artifacts} & XGBClassfier & .94 &  .83 \\
\textit{Entropy ,Peak to Mean} &  &  & \\
\hline

\end{tabular}
\label{tab:classifier}
\end{table}



\subsection{Explainable AI the in Assessment Module}

The black box nature 
of deep learning represents a challenge for analyzing their decision-making process~\cite{groza2021agents}, an issue which can be addressed by \textit{Explainable AI (XAI)}. 

\begin{figure}
    \centering
    \includegraphics[height=4cm]{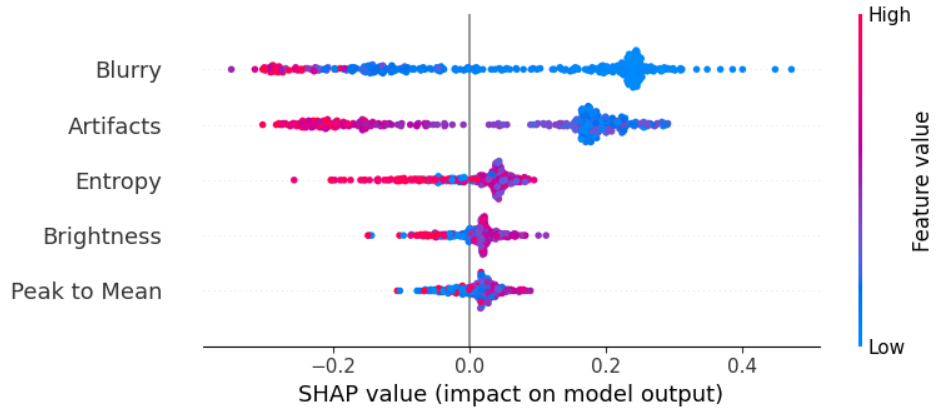}
\caption{Decision-making of binary classifier using SHAP}
\label{fig}
\end{figure}

Because out of the two most accurate classifiers, Random
Forest can be considered more interpretable, its decision-making process is analyzed using \textit{SHapley Additive exPlanations} (SHAP) (see Figure~\ref{fig}).
A \textit{positive} SHAP value pushes the prediction towards good quality, while a \textit{negative} SHAP values pushes the prediction towards a bad quality image. At the same time, a pink value represents a high feature value, while blue represent a low one.
Therefore,
\begin{itemize}
    \item \textit{Blurriness (Blurry)} - low values push towards good quality, while high values push towards a bad quality; so, high percentage of blurriness indicates a bad quality image and vice versa, which is correct
    \item \textit{Absence of artifacts (Artifacts)} - high values push towards a bad quality prediction, while lower values are more prone to indicate a good quality image; so, a bad quality image has no artifacts displayed, which is true according to theory
    \item \textit{Entropy} - A good part of mid-to-high values (purple-pink) tend to suggest a good quality image, while low values, along with very high values, are more likable to indicate bad quality; so low entropy indicates a photo of worse quality, which checks the theory; at the same time, photos which offer too much information can be classified as bad ones, suggesting they may indicate noise
    \item \textit{Brightness} - very high and low values indicate bad quality, while mid-range ones predict a good quality; this corresponds to reality as over/underexposed images are generally unsuitable for analysis
    \item \textit{Peak to Mean} - even though it helps in improving the accuracy, a conclusion regarding how the possible values contribute to the quality score cannot be determined.
\end{itemize}
\vspace{0.5cm}

\begin{figure}
    \centering
    \includegraphics[width=0.3\textwidth]{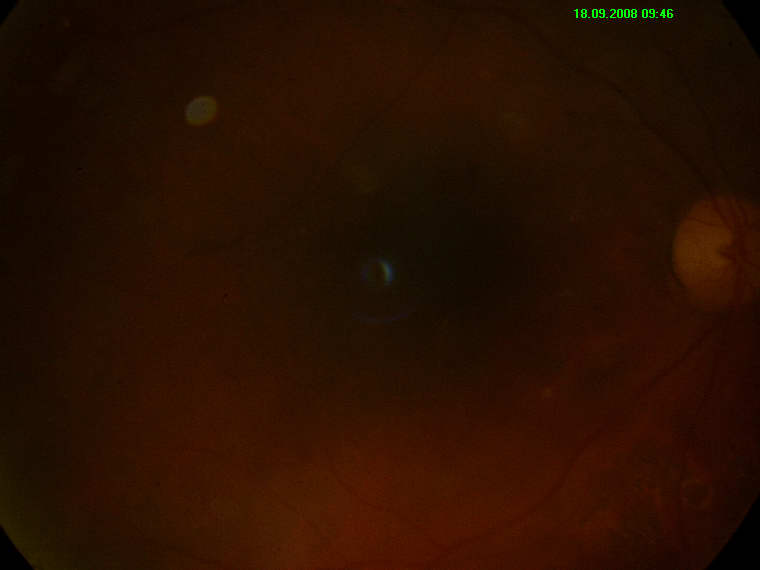}
\caption{Bad Quality image}
\label{figbad}
\end{figure}
These observations are reflected in the decision-making process of assessing individual images.
Let the example in Figure~\ref{figbad}: 
where the assessment "bad quality" image is explained by: high value of \textit{"Blurry"}, 
low value of \textit{"Entropy"}, 
high value of \textit{"Peak to Mean"}, and 
low value of \textit{"Brightness"}.

\subsection{Detecting Lesions}
Due to the sparse nature of these lesions and the majority of background pixels, average of Dice Similarity Coefficients (DSC) on the images of the dataset was used as the primary evaluation metric instead of accuracy.

Three models were initially trained for detection of EXs for 10 epochs each on the initial dataset, without enhancement methods applied and using combination of Binary Cross Entropy and Dice Loss functions, to assess how well they capture general retinal information. As presented in Table~\ref{tab:HEInitial}, following the fact that DeepLabV3+ had best DSC on validation set, the model training was extended on 50 epochs, to learn more detailed features, resulting in DSC of 0.61 on validation set and DSC of 0.58 on test set.
Additionally, post-processing techniques were applied to filter out background regions.

Table~\ref{tab:HEInitial} presents the comparison between the results obtained after variations of the training of DeepLabV3+ model. Here it can be observed that training with Dice Loss leads to a better DSC score than with Focal+ Tversky Loss. Additionally, post-processing that eliminates background lesion regions results in a higher DSC score, indicating better agreement between the automated detection and the ground truth.
\begin{table}
\centering
\caption{Detecting Hard Exudates (EX) on the initial dataset}
\begin{tabular}{l l l l}
Model &  Epochs & Loss & DSC \\
\hline
DeepLabV3 & 10  & BCE + Dice & 0.56 (val set)  \\
DeepLabV3+ & 10  & BCE + Dice & 0.57 (val set) \\
Unet++ & 10  & BCE + Dice & 0.53 (val set) \\
DeepLabV3+ & 50  & BCE + Dice & 0.58   \\
DeepLabV3+ & 50  & Focal + Tversky & 0.56 \\
DeepLabV3+, post-processing & 50  & BCE + Dice & 0.63 \\
\hline
\end{tabular}
\label{tab:HEInitial}
\end{table}

\subsection{Impact of curated, enhanced images on segmentation}


An additional setting was considered in which the initial dataset was first curated using the quality assessment classifier, after which the retained images were subjected to the proposed enhancement procedure.

Initially, a model for EXs was trained (without post-processing). Better results than on the initial dataset are obtained: 0.59 DSC in comparison with 0.58 DSC on the test set.
The approach was repeated for all lesion types, including post-processing (with the exception of SEs as they are more cohesive).  The models for EXs and SEs were trained for 50 epochs as they more visually distinct and the risk of overfitting is low, while for HAs and MAs models trained for 30 epochs were used as these lesions are more subtle and easier to misclassify due to overfitting, generalization being prioritized. For most types better results are obtained, the biggest improvement being observed in the case of HAs, while for SEs no improvement is observed (see Table~\ref{tab:Impact}).

\begin{table}
\centering
\caption{Impact of curating and enhancing the dataset  on test DSC for each lesion type using DeepLabV3+}
\begin{tabular}{lccc}
Lesion & DSC initial set & DSC enhanced set  \\
\hline
EX & .63 & .64   \\
HA & .41 & .60 \\
SE & .69 & .68 \\
MA (Patch) & .37 & .38 \\
\hline
\end{tabular}
\label{tab:Impact}
\end{table}

These results indicate that dataset curation positively influences model training, enhancing performance for the majority of the lesion types.



\begin{figure}
    \centering
    \includegraphics[height=4.3cm]{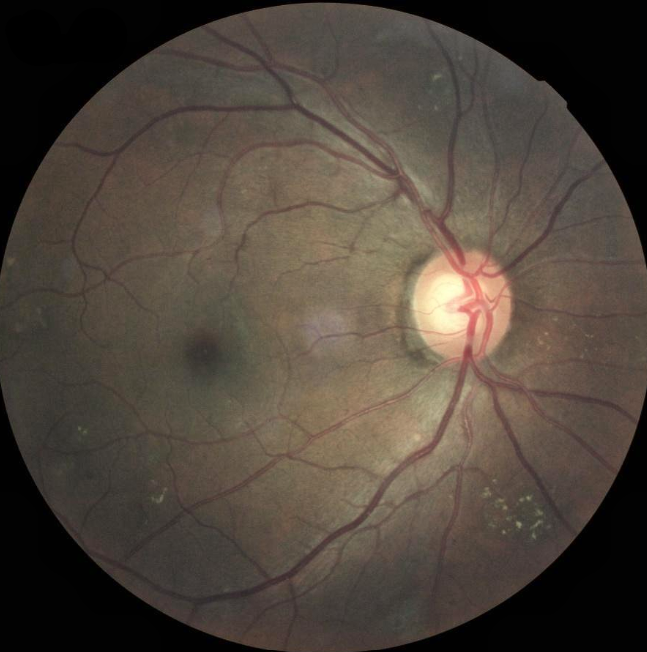}
\caption{A fundus image with lesions}
\label{lesion}
\end{figure}

For qualitative analysis, a comparison between the predicted and ground truth regions is presented for the image in Figure~\ref{lesion}.

\begin{figure}
  \centering
  \subfloat[Initial image]{%
    \includegraphics[width=0.215\textwidth]{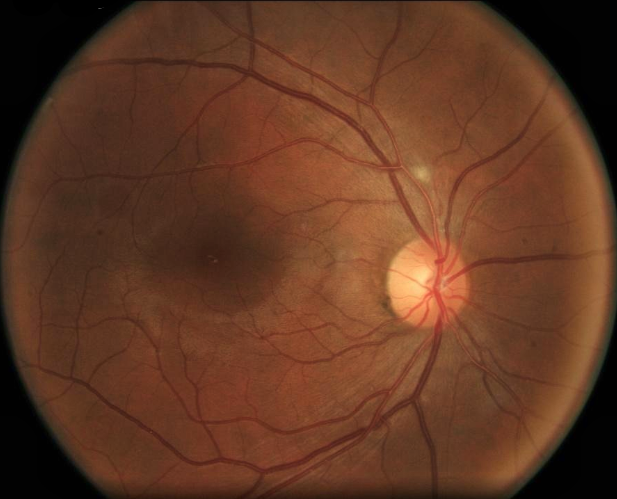}
    \label{fig:imageCWS}
  }\hfill
  \subfloat[Prediction of Soft Exudate]{%
    \includegraphics[width=0.22\textwidth]{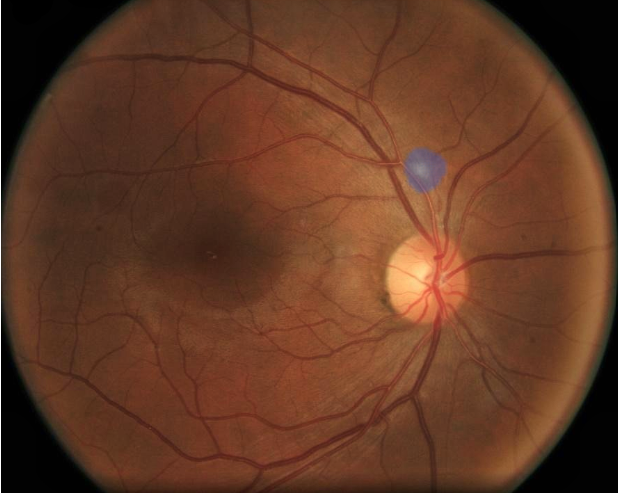}
    \label{fig:imageCWS2}
  }
  \caption{Soft Exudate (SE) Detection}
  \label{fig:CWS}
\end{figure}

\begin{figure*}
  \centering
  \subfloat[Automated annotation]{%
    \includegraphics[width=0.22\textwidth]{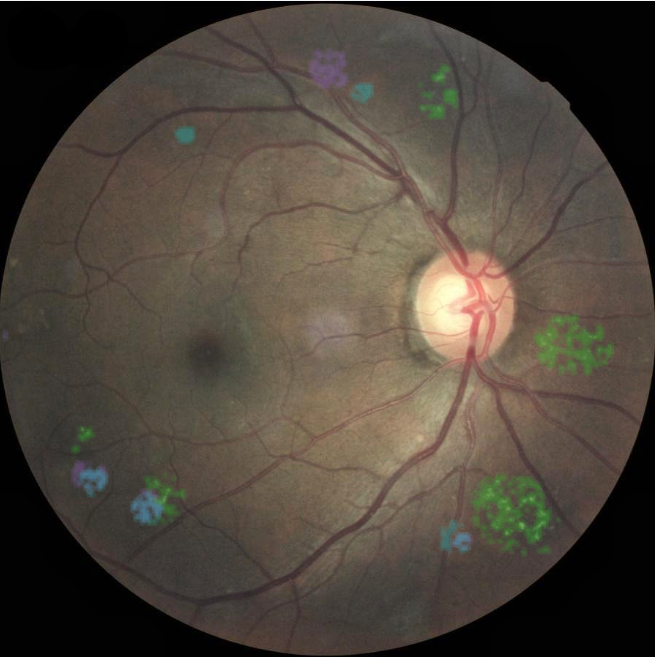}
    \label{fig:AT}
  }\hfill
  \subfloat[Ground truth - HE and HA]{%
    \includegraphics[width=0.22\textwidth]{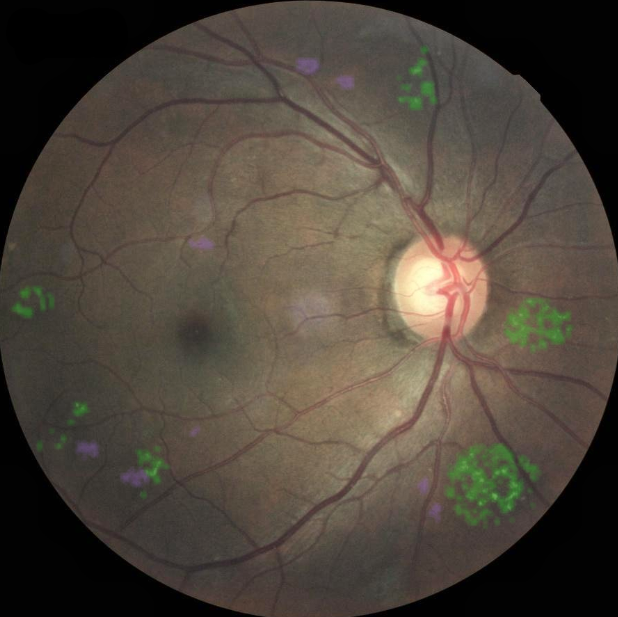}
    \label{fig:GT}
  }\hfill
  \subfloat[Ground truth - HE and MA]{%
    \includegraphics[width=0.22\textwidth]{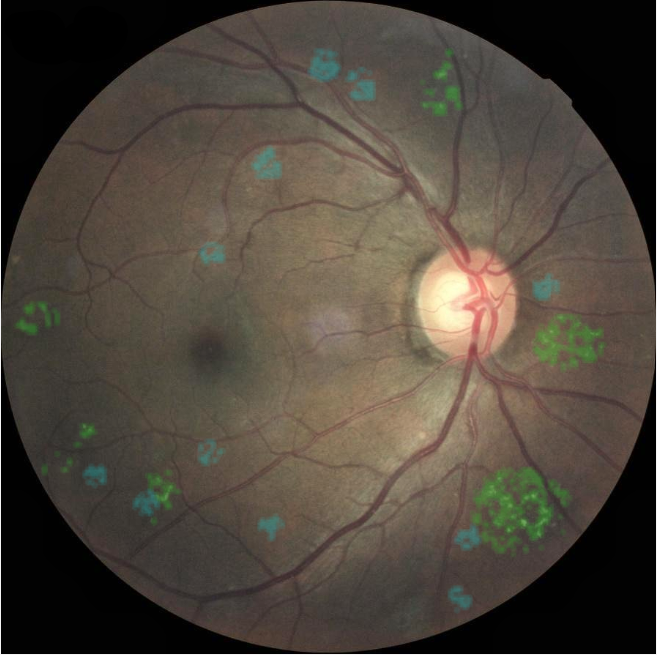}
    \label{fig:GT2}
  }
  \caption{Hard Exudate (HE - Green), Haemmorhage (HA - Purple) and Microaneurysm (MA - Light Blue) Detection}
  \label{fig:All}
\end{figure*}

In Figure~\ref{fig:All}, even in the groundtruth, the HAs and MAs are overlapping. In addition, only one region with EXs is not detected, with a part of the missed HAs are detected as MAs and only the very small HAs/MAs are not detected.
Both annotations do not consider the existence of SEs.
Another example with SE is presented in Figure~\ref{fig:CWS}, where both the automated annotation and grountruth suggest the same region.

\subsection{Inter-annotator agreement}

The framework was evaluated by two ophthalmologists with different levels of expertise -  one expert and one resident -  during the manual annotation process. The evaluation demonstrated that the application facilitated the annotation task.
We assigned a normalized expertise level of 1.0 to the specialist, and 0.6 to the resident, reflecting their relative clinical experience and familiarity with diabetic retinopathy lesions. These values were used as weighting factors in the calculation of annotation agreement scores.

The assessment is rule-based and robust, relying on established Cohen's kappa values. Considering that these formulas are sensitive to pixel-level discrepancies, especially in the case of very small lesions, a Cohen’s Kappa value around 0.2 may still indicate slight agreement on a per-lesion basis. However, for large lesions and for averaged performance across images or lesion types, a threshold around 0.4 is generally interpreted as fair agreement. In addition, the scores are determined per lesion type and not per lesion due to the computational cost.

\begin{figure}
  \centering
  \subfloat[Annotations of expert]{%
    \includegraphics[width=0.215\textwidth]{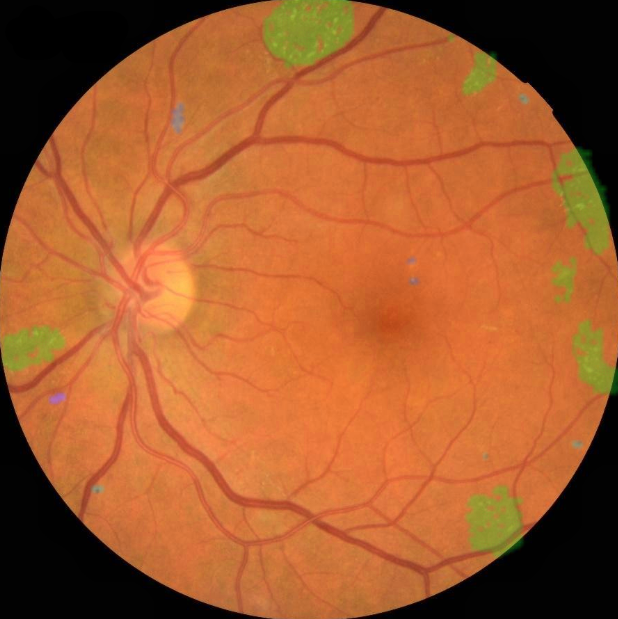}
    \label{fig:imageA1_4428}
  }\hfill
  \subfloat[Annotations of resident]{%
    \includegraphics[width=0.22\textwidth]{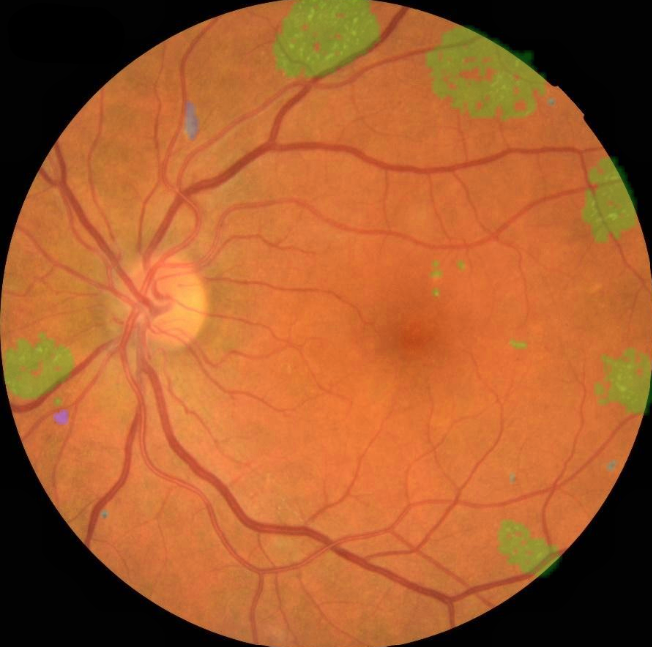}
    \label{fig:imageA2_4428}
  }\hfill
  \caption{Images with visual similar annotations: EX - Green, HA - Purple, SE - Dark blue, MA - Light blue}
  \label{fig:Annotation4428}
\end{figure}

\begin{figure}
  \centering
  \subfloat[Annotations of expert]{%
    \includegraphics[width=0.215\textwidth]{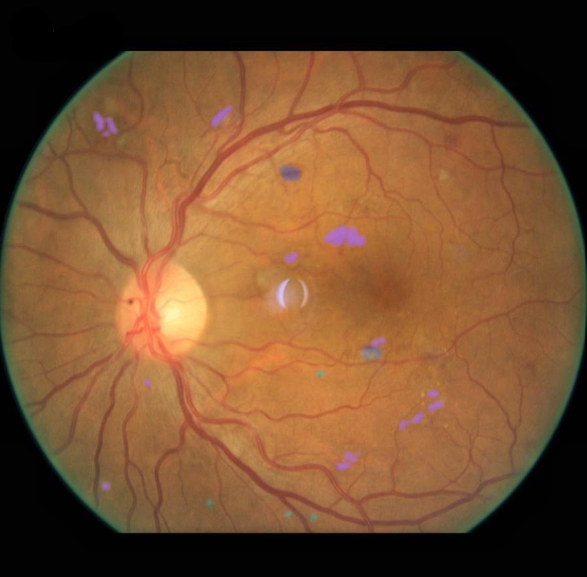}
    \label{fig:imageA1_2901}
  }\hfill
  \subfloat[Annotations of resident]{%
    \includegraphics[width=0.22\textwidth]{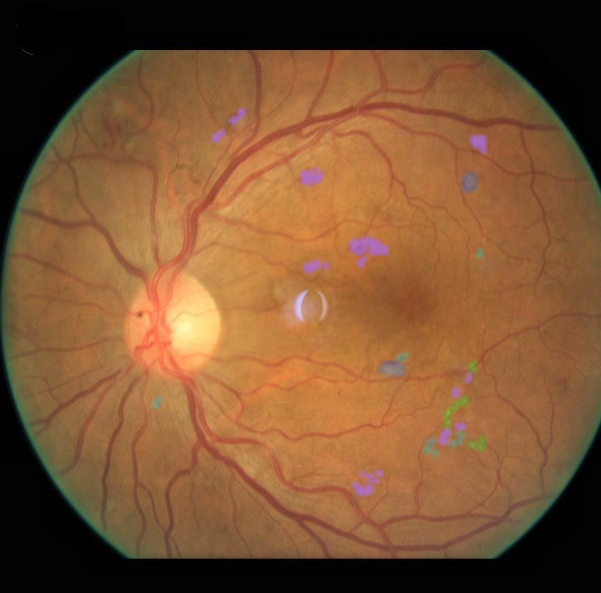}
    \label{fig:imageA2_2901}
  }\hfill
  \caption{Images with visual slightly different annotations: EX - Green, HA - Purple, SE - Dark blue, MA - Light blue}
  \label{fig:Annotation2901}
\end{figure}

Two representative examples are presented. In the first one (Figure~\ref{fig:Annotation4428}) there is visual similarity between annotations. Fair agreement is among ophthalmologists (Table \ref{tab:Agreement1}).
In the second example (Figure~\ref{fig:Annotation2901}) there are slightly more differences. It shows significant disagreement even among ophthalmologists (Table \ref{tab:Agreement2}), highlighting the need for a control mechanism to reduce annotation bias and improve consistency.

\begin{table}
\centering
\caption{Inter-annotator agreement for an image with visually similar annotations}
\begin{tabular}{lcccccc}
Lesion & Cohen Kappa & W Cohen Kappa & DSC & Weighted DSC \\
\hline
EX      & .52 & .49 & .54 & .43 \\  
HA      & .23 & .19 & .23   & .18   \\ 
MA      & .49 & .41 & .49  & .36    \\
SE      & .49 & .44 & .49  & .38  \\ 
\hline
Average & .42 & .37 & .43   & .33     \\
\end{tabular}
\label{tab:Agreement1}
\end{table}
\begin{table}
\centering
\caption{Inter-annotator agreement for an image with visually slightly different annotations} 
\begin{tabular}{lcccccc}
Lesion & Cohen Kappa & W Cohen Kappa & DSC & Weighted DSC \\
\hline
EX      & .00 & .00 & .00 & .00  \\
HA      & .31 & .26 & .31  & .24    \\
MA      & .00 & .00 & .00   & .00      \\
SE      & .38 & .31 & .38  & .28   \\
\hline
Average & .17 & .14 & .17   & .13    \\
\end{tabular}
\label{tab:Agreement2}
\end{table}

As seen in both Tables \ref{tab:Agreement1} and \ref{tab:Agreement2}, the weighted scores are consistently lower than the unweighted ones, indicating that the weighting scheme serves as a stricter form of evaluation by incorporating annotator confidence and expertise.

\section{Conclusion}

We presented a framework for curating fundus imaging data by jointly considering overall image quality and the consistency of lesion annotations, with the aim of producing reliable training corpora for AI-based detection of biomarkers associated with the presence and severity of diabetic retinopathy. The proposed pipeline integrates an explainable image-quality classifier with deep learning–driven lesion suggestions and visualization enhancement (CLAHE, gamma-based illumination correction), complemented by lesion-specific post-processing. Experiments indicate that these components can support specialists during labeling by improving lesion visibility and reducing the manual effort required for precise annotation. Importantly, both the quality classifier and the lesion models were trained using diverse and representative datasets, rather than narrowly curated sources, which improves robustness and facilitates deployment on clinician-acquired, real-world images. 
Overall, the framework promotes the development of more generalizable computer-assisted systems for diabetic retinopathy screening and diagnosis by better aligning model training with clinical data characteristics.

\textbf{Code availability}: \url{https://github.com/ancamihai/Explainable_Fundus_Image_Framework}

\textbf{Acknowledgment}: A. Groza is supported by the ”Romanian Hub for
Artificial Intelligence-HRIA”, Smart Growth, Digitization and
Financial Instruments Program, MySMIS no. 334906. The annotation was supported by University of Medicine and Pharmacy "Iuliu Hațeganu" Cluj-Napoca, with special thanks to Dr. Ioana Ințe for her valuable feedback.

\bibliographystyle{IEEEtran}
\bibliography{references}

\end{document}